\documentclass[runningheads]{llncs}
\usepackage{graphicx}
\usepackage{mathrsfs}
\usepackage{amsfonts}
\usepackage{amsmath}
\usepackage{bm}
\usepackage{multirow}
\usepackage[table,xcdraw]{xcolor}
\usepackage{color}
\usepackage{ulem}
\usepackage[misc]{ifsym} 

\begin{document}

\title{Statistical Dependency Guided Contrastive Learning for Multiple Labeling in Prenatal Ultrasound}

\titlerunning{Statistical Dependency Guided CL for Multiple Labeling in Prenatal US}


\author{Shuangchi He\inst{1,2,3}\thanks{Shuangchi He and Zehui Lin contribute equally to this work.} \and
    Zehui Lin\textsuperscript{1,2,3$\star$} \and
    Xin Yang\inst{1,2,3} \and
    Chaoyu Chen\inst{1,2,3} \and
    Jian Wang\inst{1,2,3} \and
    Xue Shuang\inst{1,2,3} \and
    Ziwei Deng\inst{1,2,3} \and
    Qin Liu\inst{1,2,3} \and
    Yan Cao\inst{1,2,3} \and
    Xiduo Lu\inst{1,2,3} \and
    Ruobing Huang\inst{1,2,3}\and
    Nishant Ravikumar\inst{4,5} \and
    Alejandro Frangi\inst{1,4,5,6} \and
    Yuanji Zhang\inst{7} \and
    Yi Xiong\inst{7} \and
    Dong Ni\inst{1,2,3}\textsuperscript{(\Letter)}}

\authorrunning{S. He et al.}

\institute{
  \textsuperscript{$1$} National-Regional Key Technology Engineering Laboratory for Medical Ultrasound, School of Biomedical Engineering, Health Science Center, Shenzhen University, China\\
  \email{nidong@szu.edu.cn} \\
  \textsuperscript{$2$} Medical Ultrasound Image Computing (MUSIC) Lab, Shenzhen University, China\\
  \textsuperscript{$3$} Marshall Laboratory of Biomedical Engineering, Shenzhen University, China\\
  \textsuperscript{$4$} Centre for Computational Imaging and Simulation Technologies in Biomedicine (CISTIB), University of Leeds, UK\\ 
  \textsuperscript{$5$} Leeds Institute of Cardiovascular and Metabolic Medicine, University of Leeds, UK\\
  \textsuperscript{$6$} Medical Imaging Research Center (MIRC), KU Leuven, Leuven, Belgium\\
  \textsuperscript{$7$} Department of Ultrasound, Luohu People's Hospital, Shenzhen, China\\}

\maketitle  

\begin{abstract}
    Standard plane recognition plays an important role in prenatal ultrasound (US) screening.
    Automatically recognizing the standard plane along with the corresponding anatomical structures in US image can not only facilitate US image interpretation but also improve diagnostic efficiency. In this study, we build a novel multi-label learning (MLL) scheme to identify multiple standard planes and corresponding anatomical structures of fetus simultaneously. Our contribution is three-fold. First, we represent the class correlation by word embeddings to capture the fine-grained semantic and latent statistical concurrency. Second, we equip the MLL with a graph convolutional network to explore the inner and outer relationship among categories. Third, we propose a novel cluster relabel-based contrastive learning algorithm to encourage the divergence among ambiguous classes. Extensive validation was performed on our large in-house dataset. Our approach reports the highest accuracy as 90.25$\%$ for standard planes labeling, 85.59$\%$ for planes and structures labeling and mAP as 94.63$\%$. The proposed MLL scheme provides a novel perspective for standard plane recognition and can be easily extended to other medical image classification tasks.
\end{abstract}

\section{Introduction}

\begin{figure}
    \centering
    \includegraphics[width=0.6\textwidth]{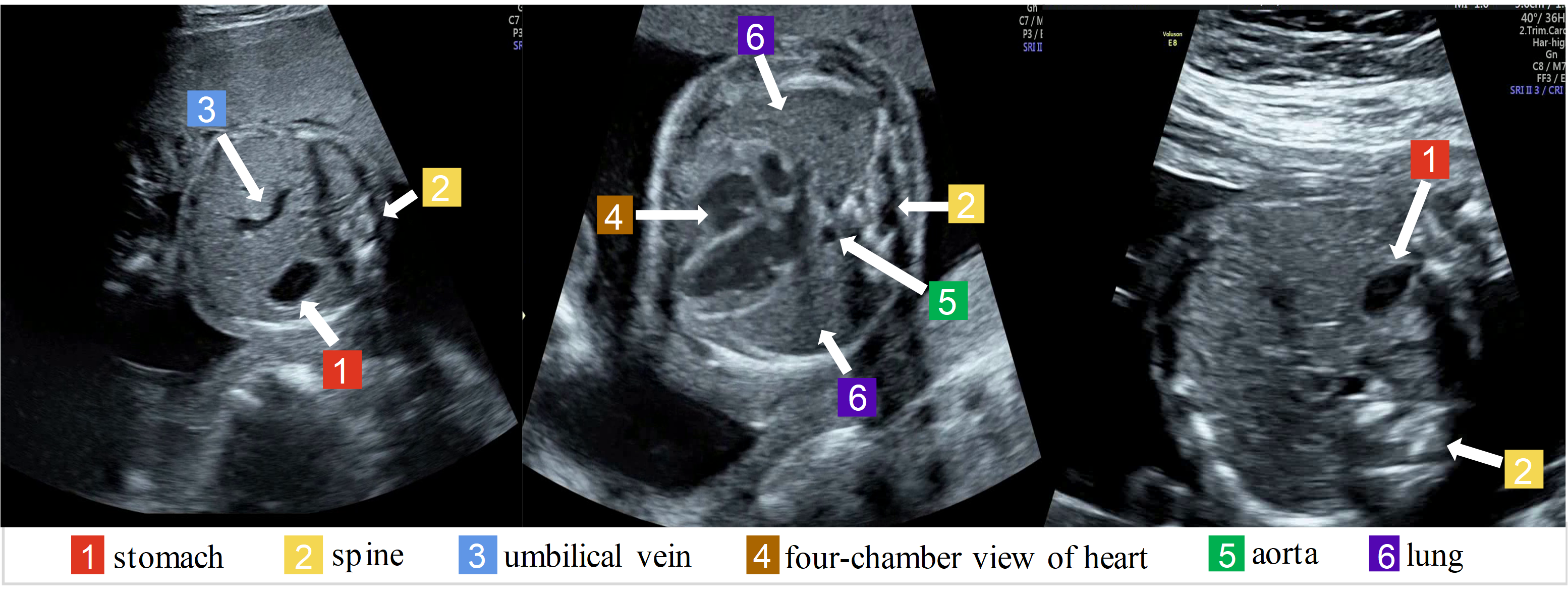}
    \caption{Left: standard plane of fetal abdomen; Middle: standard plane of four chambers of fetal heart; Right: non-standard plane around fetal abdomen. All images are annotated with multiple anatomical structure labels.}
    \label{abdomen_heart_show}
\end{figure}

Ultrasound (US) is widely used for the evaluation of fetal growth and congenital malformations in routine obstetric examinations~\cite{salomon2011practice}. During the scanning, US standard planes (SPs) that contain key anatomical structures (ASs) are selected and subsequent biometric measurements are performed~\cite{chen2017ultrasound}.
For example, the abdominal circumference (AC) is measured on the transverse plane of the fetal abdomen with umbilical vein at the level of the portal sinus and stomach bubble visible (Fig.~\ref{abdomen_heart_show}). The value of  AC is then used to estimate the pre-birth weight of a fetus~\cite{salomon2011practice,chen2017ultrasound}. In clinical practice, the standard plane (SP) selection based on ASs identification is experience-dependent, cumbersome, and suffering from the inter-observer and intra-observer variability~\cite{burgos2020evaluation}. Hence, automatic recognition of SP is desired to improve the examinations. \par

In recent years, deep learning-based methods have witnessed significant grow-th in automated SP recognition. Chen et al.~\cite{chen2017ultrasound} proposed a composite neural network framework for the automatic recognition of three SPs. Burgos-Artizzu et al. \cite{burgos2020evaluation} evaluated a large set of state-of-the-art convolutional neural networks for the classification of more than 6 maternal / fetal US planes. Cai et al. \cite{cai2018sonoeyenet} presented a convolutional neural network (CNN) framework SonoEyeNet for the detection of SPs. They found that the eye movement tends to focus on the existence of ASs. These methods could distinguish the SPs from the non-standard ones directly with the plane-level labels. However, they did not explicitly incorporate the clues of key ASs,
which limited the clinical interpretability and possible guidance for novice sonographers. Lin et al. \cite{lin2019multi} focused on the detection of key ASs, providing fine-grained information of SPs. However, as shown in Fig.~\ref{abdomen_heart_show}, the presence of anatomical structure (AS) alone does not guarantee an accurate identification of the SP, as the SP is also defined by the global image appearance and subtle details~\cite{dong2019generic}.
Furthermore, the extensive annotations of each structure with bounding boxes are also labor-intensive and are difficult to obtain. Therefore, new frameworks and methods need to be devised to recognize SP and provide additional information on key ASs simultaneously. \par

In this paper, we build a novel multi-label learning (MLL) scheme to recognize multiple SPs and corresponding key ASs at the same time. Our contribution is three-fold. ($i$) Inspired by natural language processing techniques, the word embedding \cite{hinton1986learning} is introduced to model the latent concurrency and statistical dependency among different classes, including SPs and ASs. These kinds of cues prove to be strong guidance for MLL prediction. ($ii$) To further capture the topological structures in the label space of the word embeddings, graph convolutional network (GCN) \cite{chen2019multi} is explored to propagate information between multiple classes to capture the inner and outter relationship among ASs and SPs. ($iii$) To tackle the high intra-class variation and low inter-class variation of different SPs and ASs (Fig.~\ref{abdomen_heart_show}), we further devise a cluster relabel-based contrastive learning (CRC) to align the similarity and increase discrimination across different classes. We conduct extensive experiments on a large dataset which contains 9742 US images from 920 fetuses and 39 object classes (including 10 SPs and 29 ASs). Experiments prove that, the proposed MLL method can achieve promising results in classifying multiple SPs and identifying associated key ASs. \par

\section{Methodology}

\begin{figure}
    \centering
    \includegraphics[width=0.78\textwidth]{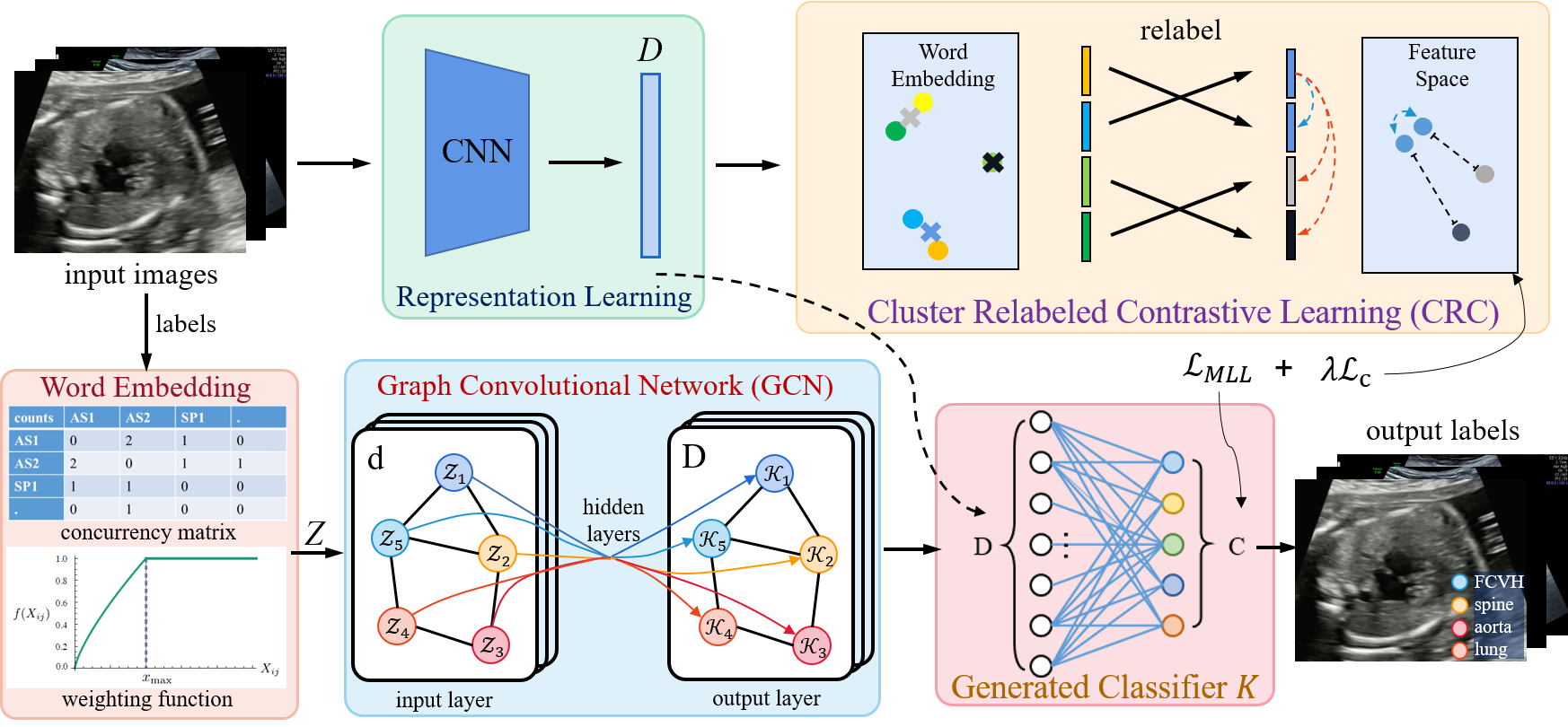}
    \caption{
    Overall framework of the proposed MLL for US image recognition. The word embeddings $Z\in\mathbb{R}^{C\times{d}}$ are generated based on the concurrency matrix and weighting function. Stacked GCNs are learned over the graph to map these word embeddings into an inter-dependent object classifier, i.e., $\bm{K}\in\mathbb{R}^{C\times{D}}$. CRC is used to improve the discrimination of the classifier. The classifier is then applied to the image representation from the input image via a CNN for MLL image recognition.} \label{framework}
\end{figure}

Fig.~\ref{framework} is the schematic view of our proposed method. We propose a MLL framework to recognize the multiple SPs and ASs simultaneously. To exploit the statistical dependency among classes, we firstly generate statistical word embeddings from label annotations. Then, we utilize GCN to model the hierarchical relationship among the classes. Further more, we propose CRC to align the high-level representation among samples of the same category. The MLL recognition output is obtained through representation learning and generated classifier. \par

\subsection{Multi-label Learning with Word Embeddings}
CNN is known for its ability in representation learning. As shown in Fig.~\ref{framework}, our MLL learning scheme is built upon a CNN to learn the feature of an image. In specific, we use ResNet \cite{he2016deep} as the backbone model. Given an input image $\bm{I}$ with a size of $448 \times 448$ pixels, we can obtain an image-level feature $\bm{x}$:
\begin{equation}
    \bm{x} = f_{cnn}(\bm{I}, \theta_{cnn})\in{\mathbb{R}^{D}},
\end{equation}
where $\theta_{cnn}$ indicates model parameters and $D=512$. \par

Inspired by the natural language processing techniques which aim to model the statistical dependency among words, phrases and sentences, we try to capture the fine-grained semantic dependency that exists among the SPs and ASs in the label space following the spirit of word embedding~\cite{hinton1986learning}. Since it is intractable to model the relationship among the SP and AS labels in prenatal US directly using the word embeddings pre-trained on natural languages, we build a corpus based on the labels from the training US dataset (An image sample is considered as a sentence, and the SP category and AS labels of the sample are considered as words.), and use the GloVe \cite{pennington2014glove} to train the word embeddings. According to the label-based sentences, we construct a concurrency matrix $X$ and use it as GloVe input. $X_{ij}$ represents the number of times class $i$ and class $j$ appear together on the same sample in the dataset. Then, the relationship between word embeddings and the co-occurrence matrix is formulated as:
\begin{equation}
    w_i^{T}\tilde{w_j}+b_i+\tilde{b_j}=log(X_{ij}),
\end{equation}
where $w\in\mathbb{R}^{C\times{d}}$ are word embeddings and $\tilde{w}\in\mathbb{R}^{C\times{d}}$ are separate context word embeddings which reduce overfitting. $b$ and $\tilde{b}$ are corresponding bias terms. \par

We can obtain the final word embeddings output $Z=w+\tilde{w}$ by optimizing the following loss function:
\begin{equation}
    J = \sum_{i,j=1}^{C}f(X_{ij})(w_i^{T}\tilde{w_j}+b_i+\tilde{b_j}-log(X_{ij}))^{2},
\end{equation}
where $C$ is the size of the vocabulary (i.e. our class number, 39), $f$ is the weighting function \cite{pennington2014glove} that adjusts the frequency of concurrency in the corpus. Word embeddings matrix $Z=\{z_i\}_{i=1}^C\in\mathbb{R}^{C\times{d}}$ hence encodes the statistical dependency and distribution relationships among different labels and can be further explored in the following sections. \par

\subsection{GCN for Class Dependency Learning}
It is important to capture the internal relationships between ASs and SPs and leverage this relationship to improve the classification performance in multi-label US image recognition.
In this paper, inspired by \cite{chen2019multi}, we explore the GCN to model the class dependency in prenatal US images, which is an effective and flexible way to capture the topological structures in the word embeddings label space represented by $Z$. Specially, GCN is built to directly map the nodes (i.e. word embeddings $Z$) of the graph into an inter-dependent classifier (Fig.~\ref{framework}).
The GCN based mapping function is defined as:
\begin{equation}
    \bm{G}^{l+1}=h(\hat{\bm{B}}\bm{G}^{l}\bm{W}^{l}),
\end{equation}
where $\bm{G}^{l}\in{\mathbb{R}^{C\times{d}}}$ are feature descriptions ($C$ denotes the number of nodes and
$d$ indicates the dimensionality of node feature) and $\hat{\bm{B}}\in{\mathbb{R}^{C\times{C}}}$ is the normalized version of correlation matrix,
and $h(\cdot)$ denotes a non-linear operation.
In every back-propagation, the transformation matrix $\bm{W}^{l}\in{\mathbb{R}^{d\times{d^{'}}}}$ will be updated. \par

As shown in Fig.~\ref{framework}, for the first layer of the stacked GCNs, the input is the word embeddings matrix $Z$. The output of the last GCN layer is $\bm{K}\in\mathbb{R}^{C\times{D}}$, which matches the dimensionality of the image representations extracted by the CNN. $\bm{K}$ contains the class dependency and hence regularizes the CNN prediction $\bm{x}$ as the final classifier. The multi-label prediction scores can be computed by applying the learned classifier to the image representation as follows:
\begin{equation}
    \bm{\hat{y}} = \bm{K}\bm{x},
\end{equation}
where the ground truth labels of an image is represented as $\bm{y}$ with $y_i=\{0,1\}$ denoting whether label $i$ appears in the image or not. The training of the whole network uses the traditional MLL classification loss as follows:
\begin{equation}
    \mathcal{L}_{MLL} = - \frac{1}{C} \sum_{c=1}^{C}(y_{i}log(\sigma(\hat{y}_{i}))+(1-y_{i})log(1-\sigma(\hat{y}_{i}))).
\end{equation}

\subsection{Cluster Relabeled Contrastive Learning}

Borrowing the idea of supervised contrastive learning \cite{khosla2020supervised}, we propose to use contrastive learning (CL) to further increase the discriminative ability of learning.
In CL, the samples belonging to the same class are encouraged to be similar to each other, while that of the different classes are encouraged to be different in high dimensional feature space. However, this principle can not be directly applied to our multi-label circumstance. One sample may have labels overlapped with the other samples, thus it is difficult to define the positive and negative sample pairs. On the other hand, semantically related concepts in the word embeddings space are found to be naturally close to each other \cite{mikolov2013efficient}. Therefore, we propose to assign every sample a new single label based on the cluster of the word embeddings and perform supervised contrastive learning. \par

Specially, as shown in Fig.~\ref{framework}, we perform the k-means clustering algorithm in the word embeddings label space $Z\in\mathbb{R}^{C\times{d}}$. We use $C$ as the sample size and $d$ as the dimensionality to generate $N$ clusters. Each sample with original multi-label $\bm{y}$ is represented as a vector $\overline{z_p}$. It is calculated through the mean value of the $\{z_i\in\mathbb{R}^{d}|y_i=1,i=1,\cdots,C\}$. The new single label $\bm{y}^*$ with $y_i^* = \{0, 1\}$, $i\in[1, N]$ is assigned to the sample according to the nearest distance among these $N$ cluster centroids. For our multi-label task, $N$ is set to 10. \par

The contrastive loss to drive the learning of relabeled samples is defined as:
\begin{equation}
    \mathcal{L}_c = \sum_{i} \sum_{j,k\neq{i}} (\alpha(1-sim(\bm{x}_i, \bm{x}_j)) + \beta(1+sim(\bm{x}_i, \bm{x}_k))),
\end{equation}
where $\bm{x}$ is the image representation, $i$ and $j$ is the positive sample pair with same $\bm{y}^*$, $i$ and $k$ is the negative sample pairs with different $\bm{y}^*$. $sim(\bm{a},\bm{b})=\frac{\bm{a}^T\bm{b}}{\|\bm{a}\|\|\bm{b}\|}$ is the cosine similarity between two vectors $\bm{a}$ and $\bm{b}$.
$\alpha$ and $\beta$ are the hyperparameter to weight the similarity.
Since there are fewer pairs of positive samples than negative samples, we empirically set $\alpha$ to 0.75 and $\beta$ to 0.25 to balance the loss weights. The total loss of the proposed method is defined as the summation of MLL loss $\mathcal{L}_{MLL}$ and contrastive loss $\mathcal{L}_c$
\begin{equation}
    \mathcal{L} = \mathcal{L}_{MLL}+\lambda\mathcal{L}_c,
\end{equation}
where $\lambda$ is the hyperparameter to weight the contrastive loss. $\lambda$ is set to 0.1 based on the validation results. \par

\section{Experimental Results}

\textbf{Implementation Details.} Our dataset contains 9742 prenatal US images from 920 fetuses, including 10 types of SP and 29 types of AS. The gestational age ranges from 18 to 28 week. The image size was set to 448 $\times$ 448. An experienced sonographer provided the ground truth labels. The dataset was randomly split into 4331, 2643 and 2768 images in fetus level for training, validation and testing. There was no overlap of fetus among datasets. Adequate data augmentation were performed. The model was implemented in PyTorch with an RTX 2080Ti GPU. We used Adam optimizer (learning rate 0.001) to train GloVe for 256 epochs to obtain 512-dimensional word embeddings. SGD optimizer (learning rate 0.01) is used to train the model for 100 epochs to obtain the MLL classifier. \par

\noindent\textbf{Quantitative and Qualitative Analysis.} We evaluated the classification in terms of the average overall precision (OP), recall (OR), F1 (OF1) and the average per-class precision (CP), recall (CR), F1 (CF1). The mean average precision (mAP), Hamming loss (HL), the accuracy of the standard plane classification (SP\_ACC) and the multi-label classification accuracy that exactly matches the categories of all targets on the image (MLL\_ACC) were also taken into consideration. Table~\ref{table} illustrates the detailed evaluation results. \par

Ablation study was conducted to compare different methods, including MLL without GCN and CL (Single-MLL), MLL with contrastive learning (MLL-CL, the non-relabeled version of CRC), MLL with CRC (MLL-CRC), MLL with GCN (MLL-GCN) \cite{chen2019multi}, MLL with GCN and vallina contrastive learning (MLL-GCN-CL) and the full model (MLL-GCN-CRC). We also compared with state-of-the-art methods, including CNN-RNN \cite{wang2016cnn} and SRN \cite{zhu2017learning}. All the above methods were pre-trained with ImageNet. The ResNet34 served as the network backbone for Single-MLL, MLL-CL, MLL-CRC, MLL-GCN, MLL-GCN-CL and MLL-GCN-CRC. We can draw the following conclusions from the Table~\ref{table}:

\begin{table}[!ht]
	\caption{Quantitative evaluation of multi-label classification methods (in \%).}
	\label{table}
	\footnotesize
	\setlength{\tabcolsep}{0.025mm}{
	\begin{tabular}{c|c|c|c|c|c|c|c|c|c|c}
		\hline
		Method     & SP\_ACC                               & MLL\_ACC                               & mAP                                   & HL                                   & OP                                    & OR                                    & OF1                                   & CP                                    & CR                                    & CF1                                   \\ \hline
		CNN-RNN    & 80.07                                 & 76.45                                 & 83.15                                 & 3.83                                 & -                                     & -                                     & -                                     & -                                     & -                                     & -                                     \\ \hline
		SRN        & 86.95                                 & 66.17                                 & 91.74                                 & 2.15                                 & 90.13                                 & 89.63                                 & 89.88                                 & 86.81                                 & 88.40                                 & 87.60                                 \\ \hline
		\hline
        Single-MLL  & 88.26                                 & 81.04                                 & 93.75                                 & 1.64                                 & 92.00                                 & 92.80                                 & 92.40                                 & 88.84                                 & 89.61                                 & 89.22                                 \\ \hline
		MLL-CL      & 88.37                                 & 81.37                                 & 93.67                                 & 1.65                                 & 92.02                                 & 92.65                                 & 92.33                                 & 88.84                                 & 89.43                                 & 89.13                                 \\ \hline
		MLL-CRC     & 88.37                                 & 81.37                                 & 93.73                                 & 1.63                                 & 92.22                                 & 92.63                                 & 92.42                                 & 89.09                                 & 89.57                                 & 89.33                                 \\ \hline
		MLL-GCN     & 89.27                                 & 84.83                                 & 94.30                                 & 1.51                                 & 92.64                                 & 93.31                                 & 92.98                                 & 89.64                                 & 90.29                                 & 89.97                                 \\ \hline
		MLL-GCN-CL  & 90.07                                 & 85.52                                 & 94.62                                 & 1.45                                 & 92.43                                 & 94.16                                 & 93.28                                 & 89.67                                 & 91.74                                 & 90.69                                 \\ \hline
		MLL-GCN-CRC & {\color[HTML]{3531FF} \textbf{90.25}} & {\color[HTML]{3531FF} \textbf{85.59}} & {\color[HTML]{3531FF} \textbf{94.63}} & {\color[HTML]{3531FF} \textbf{1.40}} & {\color[HTML]{3531FF} \textbf{92.68}} & {\color[HTML]{3531FF} \textbf{94.42}} & {\color[HTML]{3531FF} \textbf{93.54}} & {\color[HTML]{3531FF} \textbf{89.87}} & {\color[HTML]{3531FF} \textbf{92.14}} & {\color[HTML]{3531FF} \textbf{90.99}} \\ \hline
	\end{tabular}}
\end{table}

(a) GCN significantly improves the model performance (4$\%$ in MLL\_ACC) under both CL and CRC conditions (i.e., MLL-GCN-CL vs. MLL-CL and MLL-GCN-CRC vs. MLL-CRC). It is attributed to the informative class dependency extracted from the statistical word embeddings by the GCN. A similar conclusion can be deduced through the comparison between MLL-GCN and Single-MLL. \par

(b) Comparing the MLL-GCN, MLL-GCN-CL and MLL-GCN-CRC, we can draw the conclusion that, CL can increase the discriminative ability of our method by about 0.7$\%$ in MLL\_ACC. Besides, we can observe that the CRC methods consistently give better perfomances than the CL methods. The relabeled operation in CRC incorporating the fine-grained semantic in word embeddings space further boosts the similarity alignment of CL. \par

(c) Among all the state-of-the-art methods (CNN-RNN lacks some result due to its design), the proposed full model MLL-GCN-CRC achieves the best results regarding both the SPs classification and ASs identification. The statistical knowledge via graph manner and similarity alignment in MLL contributes to the capture of class dependency. \par

\begin{figure}[!ht]
	\centering
	\includegraphics[width=0.75\textwidth]{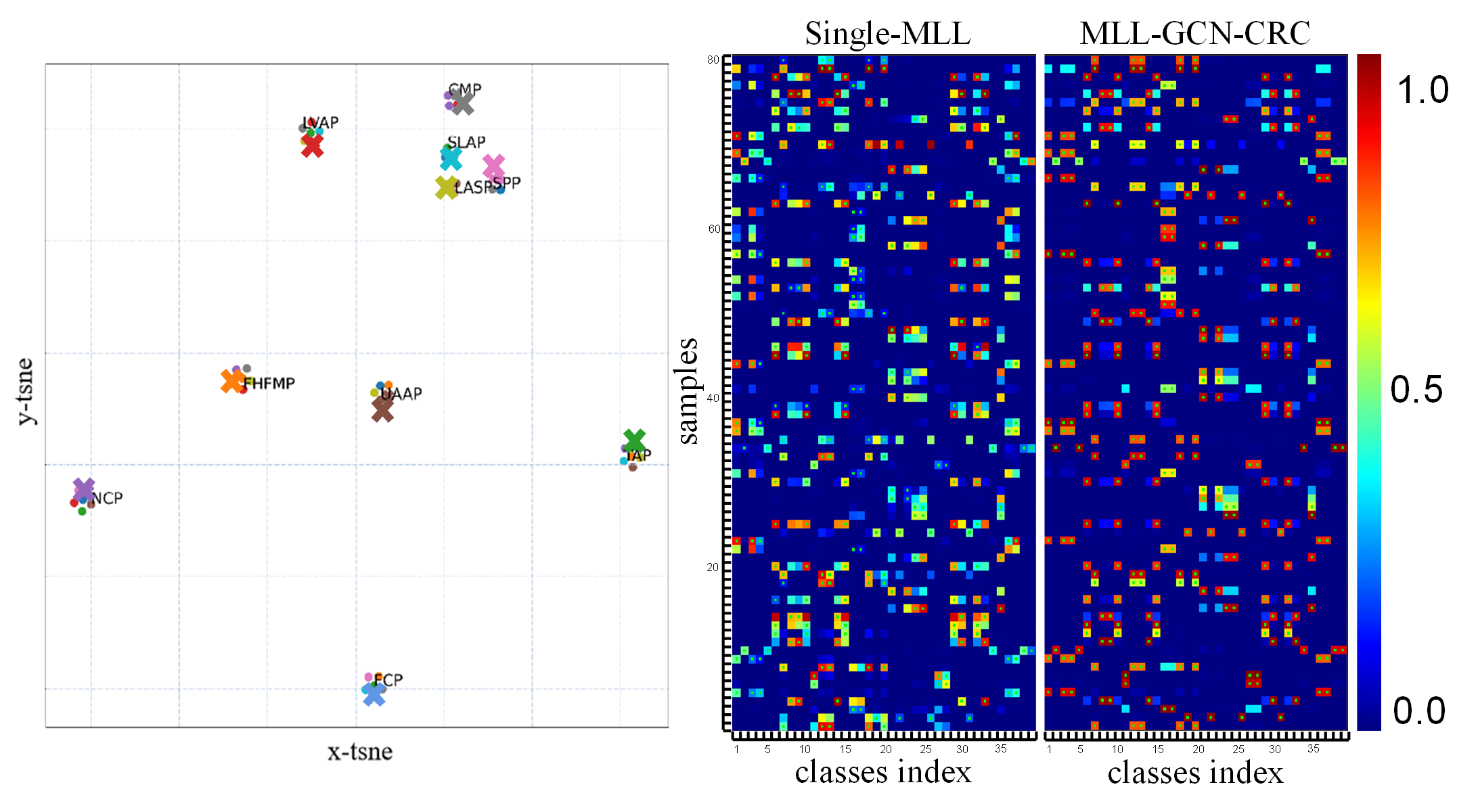}
	\caption{
		Left: the t-SNE visualization of the word embeddings space of 39 classes. The dots of different colors represent categories. The cross represents the center of cluster. Right: two score matrices of Single-MLL and our proposed MLL-GCN-CRC. Each row in the matrix represents a sample class prediction. The little green points indicate the true class and the color of matrix element indicate the predicted class scores.
	}
	\label{t_SNE-score_matrices}
\end{figure}

\begin{figure}[!ht]
	\centering
	\includegraphics[width=0.75\textwidth]{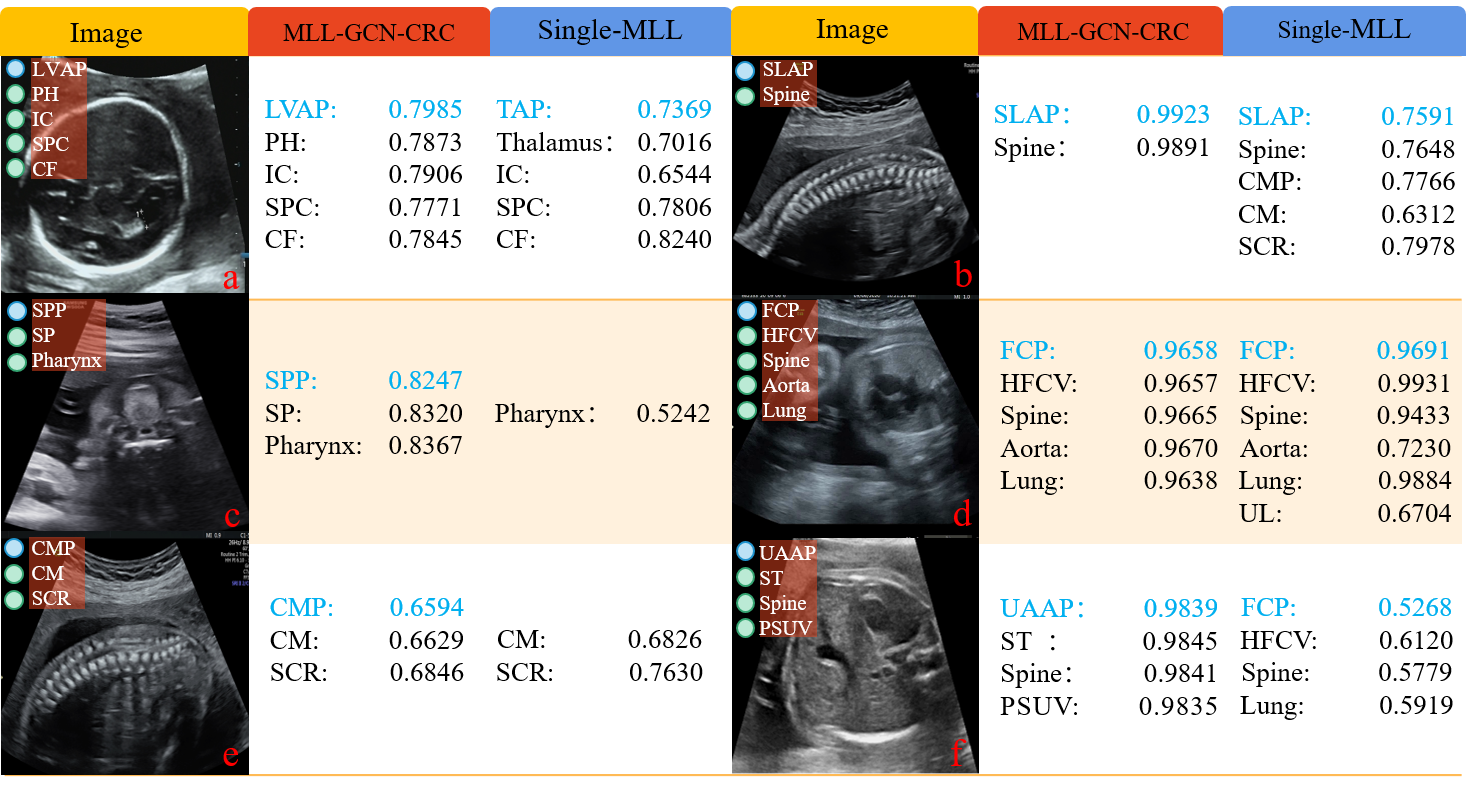}
	\caption{Typical results of multi-label recognition on fetal US. Red box for ground truth. Blue circle for SPs and green circle for ASs.}
	\label{ML_recognition_results}
\end{figure}

In Fig.~\ref{t_SNE-score_matrices}, we can observe that the related ASs and SPs embedding clustered together naturally, which builds a more semantic-reasonable label space. On the other hand, this result supports the feasibility of our CRC. The score matrices in Fig.~\ref{t_SNE-score_matrices} further illustrates the MLL prediction of some examples.
More sample score matrices can be found in Fig.~\ref{Fig-S1} in the Appendix.
It can be observed that the proposed method MLL-GCN-CRC obtains more matched cases (i.e. green point locates in the region with high score) than the Single-MLL does. This phenomenon reflects that the statistical knowledge encoded by GCN and the discriminative power enhanced by CRC are beneficial in promoting the class prediction and reducing the false positives. \par

Fig.~\ref{ML_recognition_results} shows the prediction comparisons of six samples. Among the six SPs of fetal LVAP, SPP, CMP, SLAP, FCP, and UAAP
(see the detailed name list of SP and AS in the Table~\ref{Table-S1} of Appendix)
, our MLL-GCN-CRC obtains high scores and correct predictions for most of the SPs and ASs (Fig.~\ref{ML_recognition_results} (a), (c)).
More comparison results can be found in the Fig.~\ref{Fig-S2} of Appendix.
On the contrary, the Single-MLL presents mis-classifications and false positives (Fig.~\ref{ML_recognition_results} (b), (f)). \par

\section{Conclusion}

In this paper, we propose a novel multi-label learning scheme (MLL-GCN-CRC) for multiple standard planes and corresponding anatomical structures recognition in prenatal ultrasound. Following the spirit of word embedding, the statistical concurrency knowledge is explored to capture the latent class dependency between standard planes and anatomical structures. A GCN is designed to further encode the dependency among the word embeddings. By performing relabeling based on the clusters in word embeddings space, the contrastive learning boosts the classification performance. Experiments on large dataset show that the proposed method obtains promising performances. Our proposed design is general and may inspire the community for multi-task labeling. \par

\subsubsection{Acknowledgment}
This work was supported by the SZU Top Ranking Project (No. 86000000210).

\bibliographystyle{splncs04}
\bibliography{ref}

\section*{Appendix}

\begin{table}[!ht]
    \centering
    \caption{Abbreviation and full name of structures and standard plane. In the Abbreviation
        column, blue represents the standard plane, black represents the anatomical structure, and the
        horizontal line represents no abbreviation.}
    \label{Table-S1}
    \begin{tabular}{|c|c|}
        \hline
        \textbf{Abbreviation}                & \textbf{Full name}                                    \\ \hline
        {\color[HTML]{3531FF} \textbf{SLAP}} & long axis plane of the spine                          \\ \hline
        {\color[HTML]{3531FF} \textbf{CMP}}  & plane of the conus medullary position                 \\ \hline
        {\color[HTML]{3531FF} \textbf{TAP}}  & the axial plane at the level of the thalamus          \\ \hline
        {\color[HTML]{3531FF} \textbf{LVAP}} & the axial plane at the level of the lateral ventricle \\ \hline
        {\color[HTML]{3531FF} \textbf{NCP}}  & coronal plane of the nasolabial                       \\ \hline
        {\color[HTML]{3531FF} \textbf{HFMP}} & midsagittal plane of the head and face                \\ \hline
        {\color[HTML]{3531FF} \textbf{SPP}}  & soft palate plane                                     \\ \hline
        {\color[HTML]{3531FF} \textbf{FCP}}  & four-chamber view plane                               \\ \hline
        {\color[HTML]{3531FF} \textbf{UAAP}} & upper abdominal axial plane                           \\ \hline
        {\color[HTML]{3531FF} \textbf{FLAP}} & long axis plane of the femur                          \\ \hline
        \textbf{CF}                          & cerebral falx                                         \\ \hline
        \textbf{PH}                          & posterior horn                                        \\ \hline
        \textbf{SPC}                         & cavity of septum pellucidum                           \\ \hline
        \textbf{CM}                          & conus medullaris                                      \\ \hline
        \textbf{SCR}                         & sacro-coccyx region                                   \\ \hline
        \textbf{-}                           & thalamus                                              \\ \hline
        \textbf{IC}                          & intact cranium                                        \\ \hline
        \textbf{NA}                          & apex of nose                                          \\ \hline
        \textbf{NB}                          & nasal bone                                            \\ \hline
        \textbf{-}                           & palate                                                \\ \hline
        \textbf{-}                           & mandible                                              \\ \hline
        \textbf{SP}                          & soft palate                                           \\ \hline
        \textbf{-}                           & pharynx                                               \\ \hline
        \textbf{HFCV}                        & four-chamber view of heart                            \\ \hline
        \textbf{-}                           & aorta                                                 \\ \hline
        \textbf{-}                           & lung                                                  \\ \hline
        \textbf{ST}                          & stomach                                               \\ \hline
        \textbf{PSUV}                        & umbilical vein at the level of the portal sinus       \\ \hline
        \textbf{FD}                          & femur diaphysis                                       \\ \hline
        \textbf{-}                           & spine                                                 \\ \hline
        \textbf{UL}                          & upper lip                                             \\ \hline
        \textbf{LL}                          & lower lip                                             \\ \hline
        \textbf{-}                           & chin                                                  \\ \hline
        \textbf{-}                           & nostril                                               \\ \hline
    \end{tabular}
\end{table}

\begin{figure}
    \centering
    \includegraphics[width=0.78\textwidth]{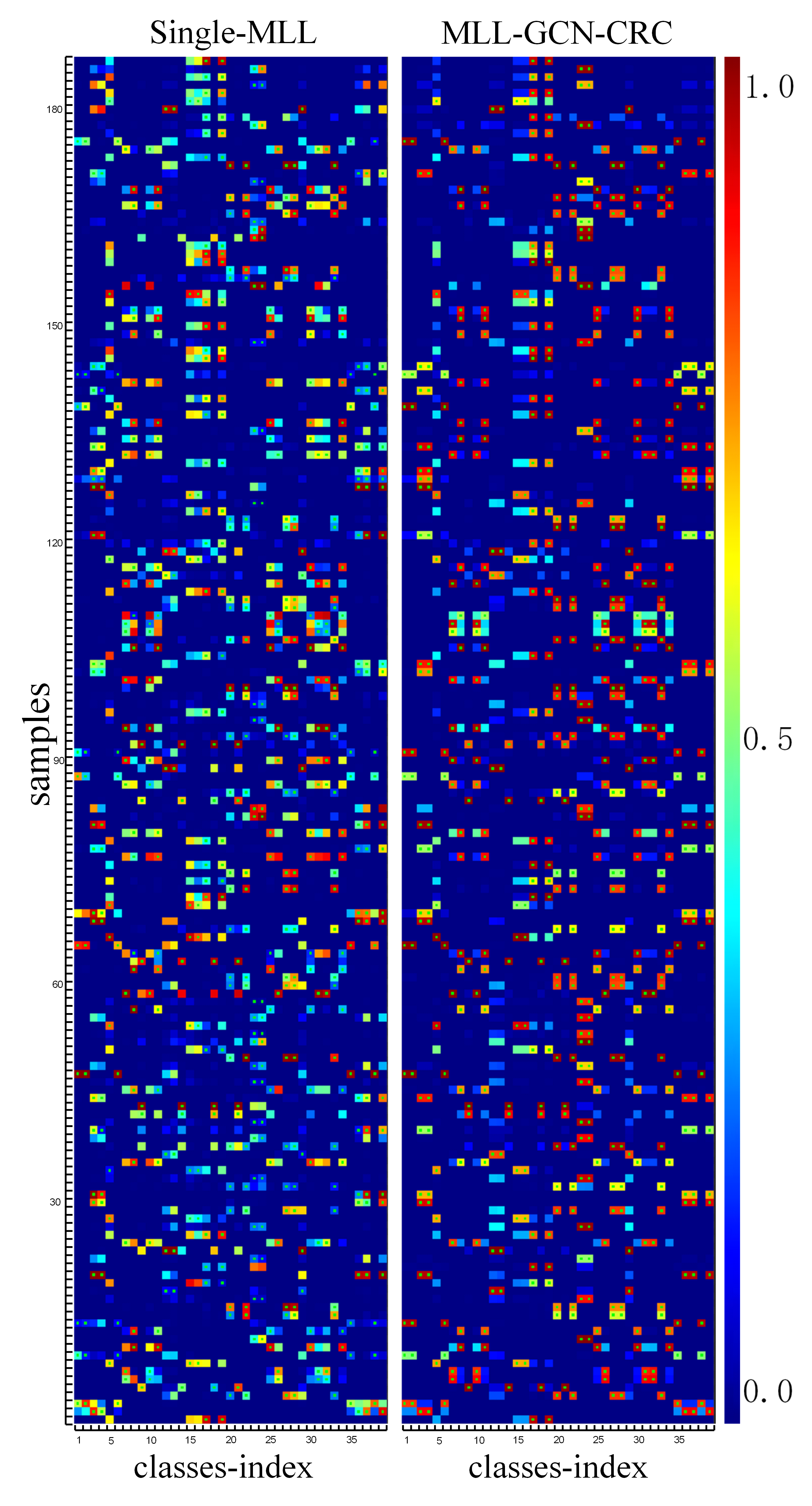}
    \caption{Two score matrices of Single-MLL and MLL-GCN-CRC. Each row in the matrix represents an example class prediction. The little green points indicate the true label and the color of matrix element indicate the model output score.}
    \label{Fig-S1}
\end{figure}

\begin{figure}
    \centering
    \includegraphics[width=0.98\textwidth]{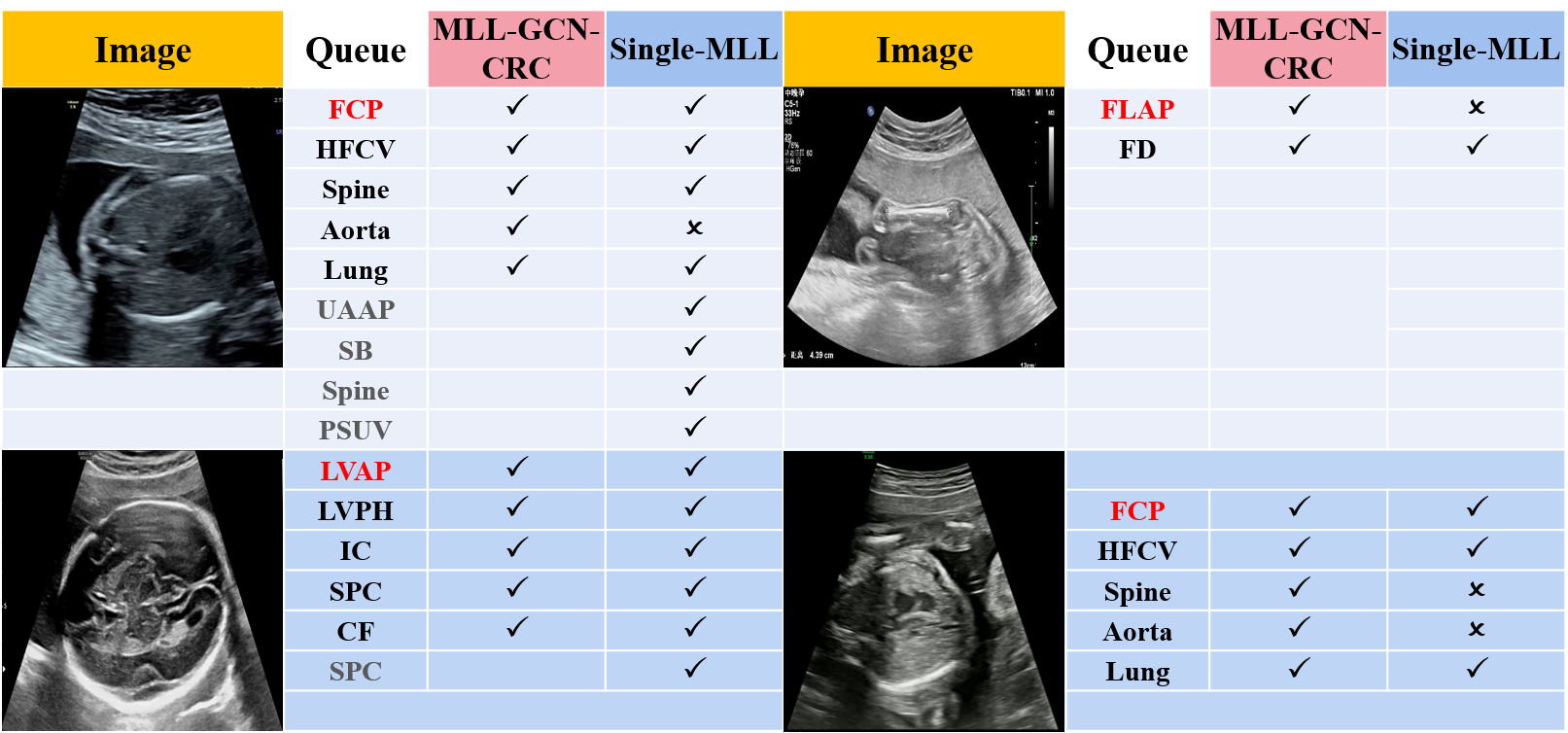}
    \caption{The result comparison between Single-MLL and MLL-GCN-CRC. In the Queue column, the bold font represents the category in the ground true, the red bold represents the standard plane category, and the gray non-bold represents the category that does not exist in the ground true. In the MLL-GCN-CRC and Single-MLL columns, a check indicates that the category is predicted, and a cross indicates that the category is not predicted.} \label{Fig-S2}
\end{figure}

\end{document}